# How Group Lives Go Well

John BEVERLEY[a,b,c] and Regina HURLEY[a,b]

[a] *National Center for Ontological Research*
[b] *University at Buffalo*
[c] *Institute for Artificial Intelligence and Data Science*

**Abstract.** This paper explores the ontological space of group well-being, proposing a framework for representing collective welfare, group functions, and long-term contributions within an ontology engineering context. Traditional well-being theories focus on individual states, often relying on hedonistic, desire-satisfaction, or objective list models. Such approaches struggle to account for cases where individual sacrifices contribute to broader social progress—a critical challenge in modeling group flourishing. To address this, the paper refines and extends the Counterfactual Account (CT) of well-being, which evaluates goodness of an event by comparing an individual's actual well-being with a hypothetical counterpart in a nearby possible world. While useful, this framework is insufficient for group-level ontologies, where well-being depends on functional persistence, institutional roles, and historical impact rather than immediate individual outcomes. Drawing on Basic Formal Ontology (BFO), the paper introduces a model in which group flourishing is evaluated in terms of group functional, where members bear roles and exhibit persistence conditions akin to biological systems or designed artifacts. This approach enables semantic interoperability for modeling longitudinal social contributions, allowing for structured reasoning about group welfare, social institutions, and group flourishing over time.

## 1. Introduction

Well-being has been explored for thousands of years [1]; it is a popular topic among researchers, with scholars from various disciplines contributing to its understanding. In recent decades, there has been a significant increase in research on well-being within behavioral and biomedical sciences, including psychiatry, personality and social psychology, narrative, and clinical psychology [2, 3, 4, 5, 6]. The study of well-being has led to the development of scales and measurement tools, such as the Short Warwick-Edinburgh Mental Well-Being Scale aimed at assessing the phenomenon [7, 8]. Moreover, the existence of specialized journals, such as the *International Journal of Qualitative Studies on Health and Well-Being* and the *International Journal of Wellbeing*, demonstrates the academic community's continued interest in this field.

Despite this interdisciplinary attention and empirical productivity, existing research frequently approaches well-being as a largely operational or univocal construct, paying scant attention to its fundamental ontology—what exactly well-being is, and how varying characterizations may interconnect. Moreover, there has been relatively little exploration into how individual well-being relates ontologically to collective forms of

well-being—which we call **group flourishing**. This omission is significant because individual well-being does not exist in isolation but within complex social contexts, suggesting that a deeper ontological understanding of individual well-being necessarily involves examining its relationship to groups.

This article addresses this gap by offering an ontological foundation for modeling both individual and collective dimensions of well-being explicitly suitable for ontology engineering tasks. Specifically, it leverages Basic Formal Ontology (BFO) to formally define roles, functions, and group entities in ways that support practical applications, such as semantic interoperability among ontologies representing complex social data. By providing clear modeling guidelines grounded in philosophical rigor but translated into ontology design considerations, the proposed framework aims to facilitate the systematic representation and analysis of group well-being and group flourishing across diverse applied ontological projects.

## 2. The Counterfactual Analysis

Suppose Jack and Jill run up a hill to fetch a pale of water as part of an exercise routine. Jack tumbles down, hits his head, and experiences pain. Jack hitting his head is a bad event for Jack. One explanation for why trades on the intuition that Jack would have been better off had he not hit his head. This intuition is often formalized as [9, 10, 11, 12, 13, 14]:

> **COUNT** The value of an event $e$ for an agent $S$ at time $t$ in world $w$ is equivalent to the well-being of $S$ at $t$ in $w$ minus the well-being of $S$'s counterpart at the nearest[1] $w'$ at $t$ where event $e$ does not occur

The badness of Jack hitting his head is simply Jack's well-being minus the well-being of his counterpart who does not hit his head. The calculation depends on what "well-being" amounts to, but **COUNT** is allegedly neutral between common positions on well-being, such as *hedonism*, *desire-satisfaction*, and *objective list theories*.

- If well-being is understood in hedonistic terms [17], i.e. as merely a balance of pleasure and pain, then Jack's well-being is arguably lower than it would have been otherwise.
- If well-being is understood along the lines of a desire-satisfaction theory [18], i.e. insofar as one satisfies one's desires, then Jack's well-being is low assuming Jack did not desire to hit his head.
- If well-being is understood in terms of an objective list theory of value [19], i.e. promoting goods among a list such as rational thinking, family, and health, then Jack's well-being is lower when he hits his head than it would have been otherwise, since his health is undermined.

---

[1] The nearest world where $e$ does not occur is determined by similarity, i.e. $w$ is similar to $w'$ when there is some property $P$ instantiated in both. More properties in common means greater similarity. Events are changes in property. **CT** is compatible with the nearest world being unique [15] or not [16].

**COUNT** provides a philosophically neutral ground on which to analyze theories of well-being.[2] However, **COUNT** being restricted to short periods raises a wrinkle [4, 12]. To illustrate, suppose we specify well-being in terms of pleasure/pain, and running up a hill is grueling for Jack and Jill. **COUNT** predicts Jill's run is bad for her, since she is in pain, and plausibly Jill's nearby counterpart who is not running, is not. But surely Jill's exercising is good for her; it prevents *future* pains. What seems missing is a link between well-being at a time and Jill's future well-being.

One contribution to the literature on this topic involves ironing out this wrinkle by extending **COUNT** to accommodate time. First, define the aggregation of well-being:

>**AGGRO** *S*'s welfare in *w* is a function of the values of events $e_1, e_2, \ldots e_n$ for S in w at corresponding times $t_1, t_2, \ldots t_n$

**AGGRO** accepts inputs from **COUNT**. We need a comparative principle of well-being over time as well:

>**WELFARE** The value of a life *p* for *S* at interval *i* in *w* is equivalent to the welfare of *S* in *w* at *i* minus the welfare of *S*'s counterpart at the nearest *w'* at *i*

**WELFARE** provides a valuation of a life by comparison to a counterpart's life. Jill's life is of higher value according to **WELFARE** if her counterpart's life has lower welfare.

*2.1. Additive Assumption*

**WELFARE** links aggregate well-being to welfare, but here we discover another wrinkle. A common assumption is that welfare is the sum of well-beings at times [21, 22, 23]. There are reasons to resist this assumption [13, 24]. Suppose Jack is a successful psychologist researching substance abuse treatment. Consider two paths Jack may have taken to studying substance abuse. In the first, after acquiring an undergraduate degree, Jack stays with his parents, reads about substance abuse, meets, and works with recovering addicts. After four years, Jack attends graduate school. In the second scenario, after undergraduate graduation, Jack does not stay with his parents, and this leads to financial difficulties, then to substance abuse, then successful recovery. After four years, Jack attends graduate school. Suppose Jack's well-being outside the four-year window is approximately the same in each scenario. According to **AGGRO**, Jack's aggregate well-being is lower in the second scenario. Then if we identify Jack's aggregate well-being with Jack's welfare, then Jack's welfare is lower too. But this is not obviously correct. Given Jack's goal of specializing in substance abuse treatment, arguably Jack's welfare in the second scenario is greater than his welfare in the first, despite the lower well-being.

    This suggests the relationship between aggregate well-being and welfare is not additive. Put another way, the value of Jack's life cannot simply amount to adding up the well-being calculations it exhibits over time; welfare seems sensitive to *well-being distributions*. The troubles Jack encountered going through substance abuse plausibly add more value to his welfare, given his goals of being a substance abuse researcher. There are several ways to accommodate. For our purposes, stipulating a positive correlation between aggregate well-being and welfare suffices, since then Jack's welfare

---

[2] **CT** appears to underwrite – in spirit if not in letter – evaluations of intervention impact [20].

in the second scenario is arguably higher than in the first.[3] Such a constraint will allow us to make the right predictions in cases similar to Jack's, even if we punt on what exactly this correlation amounts to. And *at least* correlation should be expected, since lives with more frequent, longer-lasting, positive well-being are intuitively high in welfare. We must then bridge aggregate well-being with welfare, providing a further contribution to the literature on well-being:

> **BRIDGE** *S*'s aggregate well-being is positively correlated with *S*'s welfare

Returning to Jill, we see exercise is bad for her at a time by **COUNT**, which is input to **AGGRO**. But Jill's welfare is high according to **AGGRO** and, we might presume, is positively correlated with her aggregate well-being by **BRIDGE**. It is thus in Jill's interest to exercise; no pain, no gain.

*2.2. Setting the Stage*

Call the collection of **COUNT**, **AGGRO**, **WELFARE**, and **BRIDGE** the *Counterfactual Account*. The Counterfactual Account provides an intuitive explanation for why lives are good or bad.[4] In what follows, however, I highlight a gap in the account with respect to its neutrality and with an eye towards leveraging it as a foundation for ontologies of well-being. Specifically, I argue this account cannot remain neutral between hedonism, desire-satisfaction, and objective list theories of well-being while accommodating the value of an agent's life as contributing to later *social progress* as a member of a group. After making this case, I then extend the Counterfactual Account to cover group flourishing, showing how this extension can accommodate such value while retaining the intended neutrality. Additionally, I illustrate how such accommodating results in a firm foundation for relevant ontology engineering in this space.

**3. Group Flourishing**

Suppose Jessica was a women's rights activist long before such ideas gained wider traction. Suppose Jessica spent much of her life frustrated, in psychological/physical pain, and unsatisfied since her noble activities had little effect. Consider some frustrating event for Jessica, i.e. being incarcerated for attempting to hold a rally. **COUNT** predicts this is bad for Jessica. Given the case, by **AGGRO** Jessica's aggregate well-being is low. And presumably, by **WELFARE** Jessica's welfare is low too, since – we might assume – there are nearby counterparts of Jessica's that witness more progress on women's rights issues. This aligns with **BRIDGE**, i.e. positive correlation between aggregate well-being and welfare, and results in Jessica's welfare being predicted as rather low, which seems the right prediction.

Now, suppose long after Jessica's death, her struggle is remembered by more successful activists as a milestone in the fight for equality. Arguably, Jessica sacrificed her welfare and well-being for the sake of eliminating discrimination, and this adds value to her life, even though Jessica never experienced successes.[5] It is tough to square this with implementations of the Counterfactual Account. Indeed, this seems a significant gap in both the traditional account as captured in **COUNT** and the extended account provided above.

---

[3] Positive correlation accommodates sloping well-being distributions [25].
[4] Detractors have suggested counterexamples, though largely targeting **COUNT**. For example, suppose Jack intends to surprise Jill with $100 tomorrow. Before they meet Jack changes his mind. By **COUNT**, since in the Jack's counterpart gives Jill's $100, Jack's change of heart is bad for Jill. It seems odd to say Jack's change of heart harms Jill. This is not so much a counterexample, though, as a misunderstanding of the case. Jack indeed *does* harm Jill, but it is *excused* [14].

Jessica dies before progress has been made on the issues that she cared about, and it seems implausible that Jessica's well-being and welfare might change after death.[6] Suppose we say well-being and welfare are understood in terms of balancing pleasure and pain. Since the ratio of pleasure and pain for Jessica does not change due to future progress, we seem to lack room to accommodate any additional value of Jessica's life. Suppose we say instead well-being and welfare involve desire-satisfaction, e.g. Jessica desired future activists would be more successful. Then Jessica's past desires are satisfied after death. But this suggests well-being and welfare for the dead can change, which is tough to accept [24, 26, 27]. Suppose we say well-being and welfare involve an objective list, e.g. Jessica's life exhibited morally good action, justice, etc. Then Jessica's well-being and welfare were high after all. Well-being and welfare involving an objective list, I think, gets closest to capturing what is valuable about Jessica's life. But the Counterfactual Account *is supposed to be neutral* between these three theories, and yet only the objective list version seems equipped to accommodate Jessica.

Insofar as we are engaged in the evaluation of the Counterfactual Account as a foundation for an ontological representation of well-being and welfare, we aim to provide as neutral a grounds as possible across theories of well-being. Cases like Jessica's suggest there is this a methodological gap worth addressing. We must find a way to accommodate cases like Jessica's while maintaining neutrality on theories of well-being. All is not lost. I next outline an independently motivated extension to the Counterfactual Account to cover group well-being and welfare, which has the benefit of maintaining neutrality and accommodating cases like Jessica's. The result will be a neutral foundation for ontology development concerning well-being and welfare.

*3.1. Groups and Functions*

There are persuasive arguments that groups [28] – special collections of agents - have irreducible [29] mental states [30, 31], virtues [31, 32], can be blamed [32, 33], and praised [34, 35]. Little, however, has been said about what we might call **group flourishing**, that is, how best to evaluate – in a sense – the well-being of groups over time. To make progress on this topic, a word is in order regarding what groups are ontologically speaking. At a high level, groups are not *merely* aggregates of members; groups bear some *function*, a special sort of property. Resources from the widely-used Basic Formal Ontology (BFO) [36] will help us keep sure footing as we set out to characterize group function, which will lead us to group flourishing.

Besides already including an ontological representation of function, BFO is a natural choice for ontology engineering projects. BFO was initiated to facilitate logically consistent representations of biological and medical data. Today, major users of BFO include developers in the Open Biological and Biomedical Ontologies (OBO) Foundry [37], the Industrial Ontologies Foundry (IOF) [38], and the Common Core Ontologies (CCO) ecosystem of military- and security-related ontologies [39]. Moreover, BFO is used in over 700 open-source ontology initiatives, providing a neutral starting point for ontology engineering. It is thus a natural starting point here for ontology engineering.

---

[5] One is reminded of Moses able to see but not enter the promised land.

[6] [5] calls this the problem of non-identity i.e. how can something that does not exist be harmed? We might suppose Jessica has well-being and welfare when she does not exist, but this makes it tough to distinguish between agents who do not exist and, say, shoes [6].

Generally speaking, properties can be divided into those always manifesting, e.g. shape, size, and those not always manifesting, e.g. solubility, fragility [40]. Within the context of BFO, the former are **qualities** and the latter are **dispositions**. Dispositions are, roughly, properties borne by entities that have material parts and underwrite what those bearers can do when the relevant disposition is realized. For example, a portion of $H_2O$ bears various dispositions, manifested in various environments, such as when the portion transforms into gas, liquid, or solid. Characteristic of dispositions is that if, in our example, a portion of $H_2O$ loses its disposition to, say, transform into ice, then necessarily something about its material structure changes. There is thus a tight connection between the gain and loss of dispositions and the material structure on which they depend. Even so, dispositions may be borne by a material entity without ever manifesting, such as when a soluble portion of sodium chloride never encounters water.

**Functions** are a species of disposition in BFO, as they need not always manifest and are intimately tied to the physical makeup of their bearer. Functions are associated with some purpose, e.g. a pencil is for writing, the heart pumps blood. In broad outline, a BFO function is a [40]:

> **FUN** Disposition existing in virtue of bearer's physical makeup where the bearer's physical makeup is a product of evolutionary processes or intentional design

Explicating further, it is sometimes said that a function is the *reason* or *explanation* for why its bearer exists. For example, a knife made by humans is designed to cut. Similarly, the human heart has its function due to its physical constitution, had as a consequence of evolutionary processes. The reason hearts exist is to pump blood, which is their function. Importantly, we do not say in BFO that organisms – such as humans – have functions, since that would entail there is some evolutionary or intentionality driven reason why humans exist. Parts of organisms nevertheless bear functions, even if the whole organism does not.

Even so, we maintain that groups have functions in the relevant sense. Consider a plausible group: a department hiring committee. Such a committee will be an aggregate of agents and will bear some (collective) disposition(s) – perhaps manifested in the selection of a list of potential job candidates to hire for an academic position; this may or may not manifest, e.g. members may deviate to other topics while discussing candidates and never supply a list. This disposition seems grounded in the committee's material structure, i.e. the members, and underwrites what members and the group are able to do, e.g. member voting ability, establish a chair, etc. Moreover, the reason, we might plausibly add, that the group exists is to serve a hiring purpose required by a broader university authority and motivated by a hiring need. It seems then that the hiring committee group satisfies **FUN** and thus bears a not just a disposition, but a function in the BFO sense.

There are further reasons to think groups have functions. First, reference to a group's function should partially explain why a group exists. Suppose otherwise. Two sorts of cases arise:

    **(i)**       Groups which lack functions entirely
    **(ii)**      Groups which have functions that do not explain why the group exists

Concerning **(i)**, groups are more than mere aggregates of agents. But if groups need not have any function, it is unclear how to distinguish groups from mere aggregates.

Concerning **(ii)**, suppose a non-profit group initially functions to promote literacy, but eventually loses this function, it being replaced with a money laundering function. It seems we have a group with a later function – money laundering – that does not explain why it initially existed. Hence, it is not necessary to even partially reference the group's present function to explain why it exists. In response, note if we suppose there is *one* group functioning to promote literacy, then *another* to launder money, then existence of each is partially explained by its respective function. But even if there is one group throughout, the existence of the group functioning to launder money *does* seem partially explained by the previous literacy promoting function, e.g. the group launders money more efficiently under the aegis of a non-profit. From another direction, suppose instead some group is claimed to exist to promote literacy, but in fact exists to launder money. Then it seems the group has a function to promote literacy, but this function does not explain why the group exists. But it is not obvious the group functions to promote literacy. The group was designed to *seem to* promote literacy. So, in either case it seems reference to a group's function to explain why the group exists is called for.

A second reason for thinking groups have functions is that groups, like functions, are often associated with standards of evaluation. For example, the uncorrupted non-profit above may be evaluated in terms of how well it promotes literacy. Indeed, reference to functions seems a basis for evaluating groups.

A third reason for thinking groups have functions is that group activities, like functions, are distinguishable from side-effects. For example, the function of the heart is to pump blood, which is distinct from the thumping noise the heart is able to make. Similarly, the non-profit group literacy promoting function is distinct from the side-effect of, say, increasing political awareness. This too suggests groups have functions.

A fourth reason is that group failures parallel the phenomenon of malfunctioning. The heart functions properly when pumping blood but may malfunction. Similarly, a non-profit may seek to improve literacy but perform this poorly, say, by simply handing out esoteric philosophical texts with no instruction. In these cases, the standard of evaluation meets performance, and it seems apt to say there is a malfunctioning. This suggests groups have functions.

*3.2. Roles and Value*

So, suppose groups do have functions. We can then adjust the Counterfactual Account. Starting with welfare, we first introduce a comparative claim concerning the value of a *group* insofar as it flourishes or not:

> **GRPLIFE** The value of a life *p* for group *G* at interval *i* in world *w* is the welfare of *G* in *w* at *i* minus the welfare of *G*'s counterpart in the nearest *w´* at *i*

**GRPLIFE** works much like **LIFE**. For example, the value of the life of a hiring committee over an interval is the welfare of this group minus the welfare of its nearby counterpart. We can understand welfare here in terms of how well or poorly the group functions. If the hiring committee fails to find a qualified candidate, by **GRPLIFE** the value of the life of the hiring committee is low, as should be expected for a group malfunctioning. In other words, the group is not flourishing. We might similarly try extending **COUNT** to groups:

> **GRPCOUNT** The value of an event *e* for a group *G* at time *t* in world *w* is the well-being of *G* at *t* in *w* minus the well-being of *G*'s nearby counterpart in *w′* at *t* where event *e* does not occur

But how do we understand group well-being in **GRPCOUNT**? It is not obvious the well-being of a group consists in, say, balancing pleasure and pain, desire-satisfaction, or objective list items.

Indeed, the first two options seem more applicable to *members* rather than the group. One might venture group well-being consists in balancing pleasure and pain for all members, but this is too broad. For one, agents have lives outside groups, and it seems mistaken to balance pleasures and pains for agents 'off the clock' in determining well-being for a group to which they belong. One might restrict pleasures and pains to those had by *active* members, but this entails groups that adopt members in severe chronic pain have low well-being. Perhaps restrict to *operative* members [33]? This pushes the worry back. Restricting determination to operative members would imply that the promotion of a member in great pain to operative status of the group would suddenly lower group flourishing.

One might instead venture group well-being consists in desire-satisfaction for all members. Again, it seems implausible to consider just any desires all members have, so restrict attention to desires of all active members. This is still no good. Suppose James is a member of a group whose aim is to establish white supremacist doctrines in public schools. But suppose James, while actively engaging in, say, demonstrations and rallies, does not care if the group achieves its aim. He participates because his friends are members. If James counts as a member, and desires relevant to group well-being are those had by all active members, then the group does not desire to establish white supremacist doctrines in public schools. But then if the group realizes its unpalatable function, by **GRPCOUNT** the group is neither better nor worse off, as it lacked the relevant desire because James did. This is surely the wrong result. One might respond that James is *not* an active member, but this is implausible. If the group is placed on a watch list by law enforcement as potentially dangerous, James should surely be included regardless of his apathy.

Perhaps well-being in terms of an objective list fits better with **GRPCOUNT**? That is, a group's well-being is high when it exhibits certain characteristics, e.g. effective management, fair labor distribution. This approach is promising, but appealing to an objective list reverses the issue observed above. Hedonism and desire-satisfaction focus too much on members; an objective list seems too focused on the *group*. A group that effectively achieves its goals likely exhibits characteristics one would find on an objective list for group well-being, but for this to help us accommodate Jessica's life, we must tie a group's exhibiting characteristics to activities by its members. **GRPCOUNT** thus does not seem a promising route.

But we can tie group function and member activities together by introducing another class from BFO, disjoint from dispositions but nevertheless not always manifesting when borne by an entity, i.e. **roles** [26]:

> **ROLE** Disposition agent *S* has which: exists because *S* is in an optional context *C*, is correlated with activities relevant to *C*, and *S* may gain or lose without physical change

**ROLE** is not restricted to only group contexts. Jack may acquire a barrier role by being pushed into someone's path. Alternatively, Jack might intentionally acquire a barrier role by walking in someone's path. To illustrate **ROLE**, suppose Jack is a member of the uncorrupted non-profit. This group provides a context correlated with relevant activities, e.g. educating others, handing out books. Jack inhabits this context optionally in that he could leave if he wanted. If Jack were to do so, he would not be physically changed. Hence, Jack satisfies **ROLE** and thus has a BFO role. Now, presumably events can be better or worse for members instantiating roles insofar as events affect correlated activities aimed towards realizing the group's function. But more than events, it seems activities themselves are valuable or not to members in roles:

> **ROLEWELF** The value of an activity *c* associated with role *R* for group *G*'s member *S* instantiating *R* at time *t* in world *w* is of positive degree to the extent it realizes *G*'s function at *t* in *w*, and neutral if not attempted by *S* at *t* in *w*

This characterization captures the intuition that if, say, Jack successfully teaches a child to read, this has a positive value for Jack qua group member. Importantly, **ROLEWELF** enforces degrees of positive value. If Jack fully realizes a relevant activity, thereby succeeding in achieving some function of his group, then Jack's activity has high value. If Jack does not engage in a relevant activity, the value of the activity is neutral. More frequently, agents will fall in between, having unsuccessfully attempted to fully realize a group's function. We should still count such contributions as having positive value, and that is captured by **ROLEWELF**.

We next introduce synchronic aggregate well-being of a group:

> **SYNCAGRO** *G*'s synchronic aggregate well-being in *w* at *t* is a function of values of activities $c_1,\ldots c_n$ associated with roles $R_1,\ldots R_n$ for members of *G* in *w* at *t*

Thus, at a time the well-being of a group depends on the degree to which members realize relevant activities. And we link group welfare to well-being:

> **IDT** The well-being of *G* at t in w is equivalent to *G*'s synchronic aggregate well-being at *t* in *w*

Since the well-being of a group is a function of the values of activities of members in roles, we have a close connection between members and groups. As a sanity check, observe this provides the flexibility to allow, say, a balance of pleasure or pain in group well-being. For Jack's activities could be valuable if they lead to a higher ratio of pleasure and pain. Similarly, there is room for desire-satisfaction, or for objective list items to explain why Jack's realization of group activities is good. And by introducing diachronic well-being:

**DIAGRO** *G*'s diachronic aggregate well-being in world *w* over interval *i* is a function of the synchronic aggregate well-beings $s_1, s_2,\ldots s_n$ - corresponding to times $t_1, t_2,\ldots t_n$ within *i* - for *G* in *w* at *i*

And bridging to group welfare:

**DIABRG** *G*'s diachronic aggregate well-being is positively correlated with G's welfare

We have our BFO-based foundations for an ontological account of well-being and group flourishing that remains neutral between standard theories of well-being while connecting the phenomenon to group flourishing through its members.

## 4. Ontological Foundations for Well-Being and Group Flourishing

Jessica's well-being and welfare were low, but it is plausible Jessica was a member of a group – instantiating a role with correlated activities - whose function was to eliminate discriminatory practices. Jessica's pursuance of these activities made some progress towards this aim, and that is sufficient for counting these activities as having positive value, by **ROLEWELF**. By **SYNCAGRO**, the group's synchronic aggregate well-being is positive, but low, when Jill is the sole member. With more members making substantial progress, **SYNCAGRO** entails the group will have higher synchronic aggregate well-being. By **IDT**, the early group has low well-being, and later group has higher well-being. By **DIAGRO**, the aggregate diachronic well-being for the group is high given the progress made by later generations. Lastly by **DIABRG**, this aggregate diachronic well-being is positively correlated with the group's welfare, i.e. the comparative value of the group's life according to **GRPLIFE**. Hence, while Jessica's well-being and welfare are low, her activities have positive value for the group, while activities of the later members promote further group flourishing. Importantly, this is *not* to say *Jessica's* well-being and welfare increase due to later member achievements. Rather, the well-being and welfare of the *group* is increased by her contributions. And this group plausibly outstrips the life of any particular member. That said, it is admittedly unclear to what extent. A natural thought is that a group cannot exist longer than the lives of each member. If, however, you are inclined to count groups by, say, legislative rules followed by those who are not members, then it is plausible a group might persist without membership [41, 42].

This was only one of our two masters. The Counterfactual Account extended here is intended to remain theoretically neutral among competing theories of well-being—hedonistic, desire-satisfaction, and objective list theories. This was in part accomplished by disambiguating the senses of well-being, welfare, and group flourishing while providing analyses of each for clarity. As a draft to more concretely illustrate how neutrality is preserved, consider the following ontological representation scenarios:

- Hedonistic: Pleasure and pain states of members involved in activities can be represented explicitly in BFO as **mental qualities**; the balance of pleasure and pain are informed by and inform the **process** realization of roles for group members. Roles have value based on attempted successful achievement of aims of the group that further the group's function.

- Desire-Satisfaction: Member desires related to group goals are formally represented, and satisfaction states - whether met not met – are plausibly **mental dispositions** tied to realizations in terms of **processes**. These dispositions are in turn connected to roles that value based on attempted successful achievement of aims of the group that further the group's function.
- Objective List: Specific attributes deemed objectively valuable (e.g., fairness, effectiveness, justice) are explicitly modeled as **generically dependent continuants** – copyable patterns that are often about other entities in the world - in the sense of goals or prescriptions, against which one's behavioral **processes** are measured. As behaviors are realizations of **dispositions**, generally, and dispositions are in turn connected to **roles**, value is based on attempted successful achievement of aims of the group that further the group's function within those roles.

By way of a final elaboration to solidify the approach: suppose Jessica's well-being is understood in terms of pleasure and pain. If one thinks group well-being should be understood in terms of well-being of members, fulfillment of activities bringing a balance of pleasure to pain fits nicely with the extended account. If one thinks group well-being and member well-being are not so tightly connected, then group well-being might be understood instrumentally. Suppose instead, Jessica's well-being involves desire-satisfaction. If group well-being is determined by member well-being, plausibly agents will desire to achieve group-related activities and when they do, the group is better off. If one thinks group well-being need not be understood in terms of desire-satisfaction of members, another story is needed to explain group well-being, but there is nevertheless room for desire-satisfaction explaining agent well-being. Suppose lastly, well-being involves objective list items. Here, whether or not group well-being is a function of agent well-being there is room to relate an objective list for agents to one for groups, assuming they are distinct. In sum, the Counterfactual Account extension accommodates the value of Jessica's life while remaining neutral between these competing theories and thus providing a defensible foundation for ontological representations of well-being, welfare, and group flourishing. We have met our masters.

## 5. Conclusion

The central aim of this paper was to address the limitations of existing philosophical analyses of well-being when applied to individuals and groups in the interest of offering a foundation for ontological representations of such phenomena, leveraging resources from BFO. BFO roles and functions provided a strategy for connecting the well-being of individuals to group flourishing, while remaining philosophically neutral yet practically useful for ontology engineering.

Jessica's case exemplified the limitations of traditional counterfactual accounts, which struggle to accommodate the broader value of individual sacrifices contributing to collective flourishing. We demonstrated how group flourishing is ontologically anchored in the function of groups and the roles individuals bear within them. Jessica, as a group member performing activities aligned with a group's function, contributes positively to the group's synchronic aggregate well-being—even if her personal well-being and welfare remain low. As such, the proposed framework captures the nuanced interplay between individual roles and group-level outcomes. Moreover, the ontological perspective defended here ensures neutrality among competing theories of individual well-being, whether interpreted hedonistically, through desire-satisfaction, or via objective lists. Individual activities within roles can be valuably represented in the ontology irrespective of the underlying theory of well-being adopted by specific researchers or applications. This framework does more than resolve a theoretical puzzle—it provides a practical, ontology-centered foundation for systematically representing complex social realities in applied ontology projects.

Next steps will involve constructing a more complete ontological model under BFO for well-being and group flourishing, perhaps by extending the top-level with more domain-specific ontologies. It will be important, for example, to clarify criteria for membership in a group that might outlive you and how it is related to information that might plausibly be about that group over time. Additionally, there is an important sense of "group" left out of discussion here, namely, as some manner of generically dependent continuant, i.e. plan or blueprint for an organization. Such a sense must be explicated for completeness of our account, but also owing to the fact that groups understood simply as aggregates of members cannot alone accommodate group persistence when there are vacancies in group structure, such as when a department chair position is not occupied. For our purposes, this may be relevant when exploring relationships between groups that existed historically and in the present day, despite there being intervening time during which the group had no members at all.